\documentclass[review]{elsarticle}

\usepackage{lineno,hyperref}

\usepackage{times}
\usepackage{epsfig}
\usepackage{graphicx}
\usepackage{amsmath}
\usepackage{amssymb}
\usepackage{multirow}
\usepackage{array}
\usepackage{booktabs}
\usepackage{subfigure}
\usepackage{algorithm}
\usepackage{algorithmicx}
\usepackage{setspace}

\journal{Pattern Recognition}

\begin{document}

\begin{frontmatter}

\title{GLAN: A Graph-based Linear Assignment Network}


\author{He Liu}
\ead{liuhe1996@bjtu.edu.cn}

\author{Tao Wang\corref{mycorrespondingauthor}}%
\cortext[mycorrespondingauthor]{Corresponding author}%
\ead{twang@bjtu.edu.cn}%

\author{Congyan Lang\corref{}}
\ead{cylang@bjtu.edu.cn}

\author{Songhe Feng\corref{}}
\ead{shfeng@bjtu.edu.cn}
\author{Yi Jin\corref{}}
\ead{yjin@bjtu.edu.cn}
\author{Yidong Li\corref{}}
\ead{ydli@bjtu.edu.cn}

\address{Beijing jiaotong University, Beijing 100044, China.}




\begin{abstract}
Differentiable solvers for the linear assignment problem (LAP) have attracted much research attention in recent years, which are usually embedded into learning frameworks as components. However, previous algorithms, with or without learning strategies, usually suffer from the degradation of the optimality with the increment of the problem size. In this paper, we propose a learnable linear assignment solver based on deep graph networks. Specifically, we first transform the cost matrix to a bipartite graph and convert the assignment task to the problem of selecting reliable edges from the constructed graph. Subsequently, a deep graph network is developed to aggregate and update the features of nodes and edges. Finally, the network predicts a label for each edge that indicates the assignment relationship. The experimental results on a synthetic dataset reveal that our method outperforms state-of-the-art baselines and achieves consistently high accuracy with the increment of the problem size. Furthermore, we also embed the proposed solver, in comparison with state-of-the-art baseline solvers, into a popular multi-object tracking (MOT) framework to train the tracker in an end-to-end manner. The experimental results on MOT benchmarks illustrate that the proposed LAP solver improves the tracker by the largest margin.
\end{abstract}


\end{frontmatter}


\section{Introduction}\label{sec:intor}
%
%
%
%
As a fundamental problem in mathematics, linear assignment~\cite{LAP} aims to assign $n$ jobs to $n$ agents under one-to-one matching constraints, meanwhile achieves the best state that the cost involved in the problem is minimum or the profit is maximum. It is closely related to many computer vision problems including point matching~\cite{lian2020concave}, handwritten mathematical expressions recognition~\cite{hirata2015matching}, and multi-object tracking~\cite{DAN,deepmot}. A well-known method for linear assignment is the Hungarian algorithm~\cite{hungarian,munkres} which can obtain the optimal solution without an exhaustive search. However, its computational complexity is extremely sensitive to the size of the problem. Consequently, using more elaborate heuristic or greedy strategies, several approximate algorithms including deep greedy switching~\cite{DGS,DGS-fast}, interior point algorithms~\cite{in_pts,dual_proj} and auction algorithm~\cite{auction} have been proposed to seek acceptable suboptimal solutions with time constraints. However, it is difficult to formulate the gradients of heuristic linear assignment solvers, which prevents them from being directly embedded into the learning frameworks.

Aiming to address this issue, the SinkHorn algorithm~\cite{sinkhorn} and its variants~\cite{sinkhorn_dis,GKHorn} are widely taken as a differentiable linear assignment layer that obtains the approximate solutions by performing normalization along both rows and columns of the cost matrix alternatively. With simple row- and column-normalization process, the SinkHorn-like algorithms often suffer from low assignment precision as the problem size increasing. Formulating the linear assignment problem as an integer linear programming task, Zeng et al.~\cite{DMMNet} proposed a differentiable matching layer that updates the assignment solution following the negative gradient direction and maps it onto the one-to-one matching constraint set using \emph{Dykstra's} algorithm~\cite{Dykstra1,Dykstra2}. However, its computational complexity is extremely sensitive to the problem size. In addition, these non-learnable approaches~\cite{sinkhorn,DMMNet} are inadaptable to fit data distributions in different tasks.

With the growing interest in using \emph{deep neural networks} (DNNs) for computer vision tasks, solving linear assignment problems using DNNs has attracted much research attention. Researchers~\cite{Deep_lap,RNN_FC,deepmot} model the assignment process with the deep learning framework based on \emph{Recurrent Neural Networks} (RNNs)~\cite{RNN} or \emph{Convolutional Neural Networks} (CNNs)~\cite{CNN}. However, these works mainly focus on representing elements in cost matrix with local or sequential features and less investigation has been devoted to utilizing the structured relationships between each pair of them. Specifically, the features captured by CNNs only collect information from local neighborhoods. Although the receptive field can be expanded through stacking several CNN layers, it always fails to cover the whole cost matrix when the size of the problem increases. The RNNs-based methods treat the cost matrix as time-sequential flow data which allows these algorithms to deal with linear assignment problems of varying sizes. However, during inference, the contribution of the previous element to the current state becomes insignificant when the sequence length expands, which may affect the performance of RNN-based models.

Addressing the above-mentioned issues, we propose a novel \textit{graph-based linear assignment network} (GLAN) aiming to improve both the accuracy and the computational efficiency. Our framework falls into the group of deep learning methods, and it is designed on top of graph neural networks. Given a cost matrix, we first transform it into a bipartite graph in which two sets of nodes correspond to the agents and jobs respectively, and edges between the two node sets are assigned with the corresponding cost values. By this way, the linear assignment problem is converted to the problem of selecting reliable edges from the bipartite graph.
Subsequently, a graph neural network is developed to perform information aggregation and updating on the constructed graph, which generates the final edge representations through several convolution operations. Finally, according to the updated graph state, our network predicts a label for each edge which indicates whether the assignment relationship is reliable or not.
Different from previous learning-based algorithms, the message passing process in the proposed framework is independent of the size and the structure of the constructed graph. Furthermore, the receptive field of the proposed network can cover the whole graph after only two-layer convolution modules. As a result, the proposed method is able to make assignment decisions efficiently for problems of varying sizes from the global scope.

To validate the advancement of the proposed method over previous approaches, we construct a synthetic dataset for LAP in which each sample contains a random cost matrix and its optimal assignment solution. The training of the network is guided under the assignment difference between the predictions and optimal solutions, as well as the one-to-one matching constraints, as the supervision signal.
We compare the proposed method with several state-of-the-art baseline methods in terms of assignment precision and computational complexity. Experimental results reveal that our method achieves the highest average assignment precision with low time consumption.
In addition, we embed all tested differentiable assignment solvers into an MOT pipeline respectively where each learning-based solver is taken as a trainable component to fit the data distribution. The tracking results evaluated under the standard MOT metrics illustrate that the proposed linear assignment solver is able to well capture the inductive information in the real scenarios and enhance the tracking performance by the largest margin on most metrics. We will release the code publically available upon the acceptance of the paper.

In summary, with the proposed learning framework for linear assignment problems, this paper makes contribution in three-fold:

(1) we convert the problem of linear assignment to learning of edge selection from a bipartite graph constructed based on the cost matrix;

(2) we propose a differentiable framework based on graph neural network to update the graph states, and the proposed model achieves excellent performance in terms of assignment precision and computation efficiency; and

(3) we embed the proposed LAP solver into a popular MOT pipeline and validate its effectiveness in boosting tracking performance.
\section{Related works}
\subsection{Heuristic Linear Assignment}
 In the 1950s, combining the graph theory and the duality of linear programming, Kuhn \emph{et al.} propose a well-known iterative method named Hungarian algorithm~\cite{hungarian} in which a maximal alternating forest is constructed in each iteration to augment the primal solution and the optimal solution can be obtained with $O(n^3)$ computation complexity.
 Due to the dynamic nature as well as the size expansion of many problems, where a solution needs to be found under tight time constraints, the heuristic algorithms whose solutions are close to the optimum have been investigated for decades.
Simulating the process of real auction, Bertsekas \emph{et al.} design the auction algorithm~\cite{auction} which is composed of two phases, the bidding phase and the assignment phase. The performance of the auction algorithm is improved by using the $\epsilon$ scaling strategy, where an $\epsilon$ is involved to update the elements in the cost (or price) matrix. Its results for linear assignment problems are close to the optimal solution, but it still suffers from the high computation complexity.
A variation of the Hungarian algorithm has been proposed in~\cite{shortest}, which starts with an initialization phase based on a naive auction algorithm and uses the shortest alternating paths to augment primal solutions.
In addition, several algorithms solve the linear assignment problems by general linear programming approaches such as the interior point method~\cite{dual_proj,in_pts} and dual forest method~\cite{dual_forest}. For example, Ramakrishnan \emph{et al.}~\cite{dual_proj} modify the Karmarkar interior-point algorithm~\cite{karmarkar} and develop an approximate dual projective algorithm for assignment problems. Akg{\"u}l \emph{et al.}~\cite{dual_forest} use the forest construction as a dual forest algorithm to solve the linear assignment problems.

To speed up the computation, researchers investigate several pseudo flow algorithms including~\cite{cost_scale} which solves the corresponding minimum cost flow problem. Besides, many heuristic approaches based on greedy strategies are devoted to seeking approximate solutions in a fast manner. In~\cite{trick-greedy}, Trick \emph{et al.} propose a greedy heuristic approach for the generalized assignment problems and prove that some randomization to the greedy approach can facilitate finding better solutions. Another similar approach is the \emph{greedy randomized adaptive search procedure} (GRASP)~\cite{grasp} where each search iteration is made up of a construction stage, where a randomized greedy solution is constructed, and a local search stage which applies local iterative improvement on the greedy solution.
Inspired by the above greedy heuristics, Naiem \emph{et al.} propose the \emph{Deep Greedy Switching} (DGS) algorithm~\cite{DGS,DGS-fast} which starts with a random initial solution and tries to find a better solution through searching within a well-defined neighborhood. These methods are significantly faster than previous heuristic methods, due to the partial greedy strategy. However, they often terminate when the objective function value reaches a local optimum.

\subsection{Differentiable Linear Assignment Solvers}
Despite the progress from the previous heuristic approaches, gradients of these algorithms are difficult to formulate using standard derivation paradigms, which limits their extension to machine learning frameworks.

Performing row-wise and column-wise normalization on the cost matrix, the SinkHorn algorithm~\cite{sinkhorn} has been widely used in learning-based frameworks as an assignment layer due to its high computation efficiency. However, its solution heavily depends on the distribution of values in the cost matrix, and always deteriorates when the size of the problem increases. Based on SinkHorn~\cite{sinkhorn}, several works~\cite{sinkhorn_dis,GKHorn} solve the linear assignment problem under the optimal transport architecture with entropic penalization~\cite{sinkhorn_dis} and greedy updating strategy~\cite{GKHorn}. These methods derived from SinkHorn~\cite{sinkhorn} achieve faster convergency in inference and better assignment performance, but still can not obtain satisfactory results on large-sized problem. Aiming to improve the assignment quality, i.e., assignment precision, Zeng \emph{et al.}~\cite{DMMNet} propose a differentiable mask matching layer, named \emph{Relax Matching} (RM), by unrolling a projected gradient descent algorithm in which Dykstra's algorithm~\cite{Dykstra1,Dykstra2} is employed to achieve the projection mapping the solution to constraint set. Based on a mathematic reasoning process, RM~\cite{DMMNet} obtains advancement in assignment precision over the SinkHorn algorithm~\cite{sinkhorn} by a large margin. But its computation complexity increases greatly with the size of problems expanding. Despite the achievements made by these approaches mentioned above, they are inadaptable to fit the distribution of data in different tasks due to their non-learnable property.

With the popularity of deep learning frameworks, several data-driven algorithms for linear assignment problems have been proposed in recent years. In ~\cite{one-layer}, the linear assignment problem has been converted to an equivalent linear continuous programming problem which is solved by a recurrent neural network without any designed parameters. However, the optimization process includes many matrix multiplication operations, making the methods inefficient to solve large-sized problems. Decomposing the $n$-to-$n$ assignment problem into $n$ multi-classification tasks, DFC~\cite{Deep_lap} and DBL~\cite{RNN_FC} address these sub-tasks with stacked \emph{fully-connection} (FC) layers and a \emph{bidirectional long short-term memory neural network} (BDLSTM) respectively. But the relationships between each pair of sub-tasks have not been taken into consideration. Regarding the cost matrix as an image with $n\times{n}$ pixels, DCNN~\cite{Deep_lap} employs the model based on CNNs to get local representations for making assignment decisions. However, the CNNs layers in their model have fixed receptive fields, which limits the generalization on problems with varying sizes.
To receive information from the global scope, two bidirectional RNNs sequentially applied to the cost matrix in a row-wise and column-wise direction in~\cite{deepmot}. However, it is not suitable to capture the global inductive information in sequential order for time-independent problems.


Our work falls into the group of deep learning algorithms for linear assignment problems.
Different from previous learning-based works, our method does not work directly on the cost matrix, but converts the problem to learning of edge selection on a constructed bipartite graph. A graph neural network is developed to perform computation on the graph for edge prediction, and its excellent performance is evaluated in both synthetic datasets and real-world MOT tasks.

\section{Problem Formulation}
\subsection{Linear Assignment Problem}
Given two sets of $n$ agents $I=\{1,2,\ldots,n\}$ and $n$ jobs $J=\{1,2,\ldots,n\}$, assigning job $j$ to agent $i$ results in a cost value of $C_{ij}$. The objective of linear assignment problems is to find the minimum total cost of assigning each job to exactly one agent and each agent to one job. The problem can be expressed as an integer binary programming problem where the decision variable $x_{ij}$ is set to $1$ if job $j$ is assigned to agent $i$, and is set to 0 otherwise. Therefore, the linear assignment problem can be formulated as:
\begin{equation}
\min{\sum_{i=1}^{n}\sum_{j=1}^{n}C_{ij}x_{ij}},
\label{LAP}
\end{equation}
subject to
\begin{equation}
\sum_{i=1}^{n}x_{ij}=1,\quad j=1,2,\ldots,n,
\label{constraint_1}
\end{equation}

\begin{equation}
\sum_{j=1}^{n}x_{ij}=1,\quad i=1,2,\ldots,n,
\label{constraint_2}
\end{equation}

\begin{equation}
x_{ij}\in \{0,1\},\quad i,j=1,2,\ldots,n,
\label{constraint_3}
\end{equation}

\subsection{Assigning with Edge Labeling}\label{subsection:edge-labeling}
Despite the progress of previous learning-based approaches~\cite{Deep_lap,deepmot,RNN_FC} that regard the cost matrix as an image or sequential data flow, the inherent structure information of the cost matrix has not been fully explored in their models.

Basically, assignments can be modeled and visualized in different ways~\cite{LAP}. In this paper, the linear assignment is converted to an edge labeling problem by constructing a bipartite graph. Specifically, given a cost matrix of $n$ size, we consider the agents and jobs as two sets of nodes in the constructed graph. Then we build edges between each pair of agent and job and assign the corresponding cost values to these edges, i.e., the initial value on the edge $(i,j)$ is assigned with $C_{ij}$. Finally, the value in edge $(i,j)$ predicted by our framework indicates the assignment between agent $i$ and job $j$ is reliable or not.
\begin{figure}
\begin{center}
\includegraphics[width=0.6\linewidth]{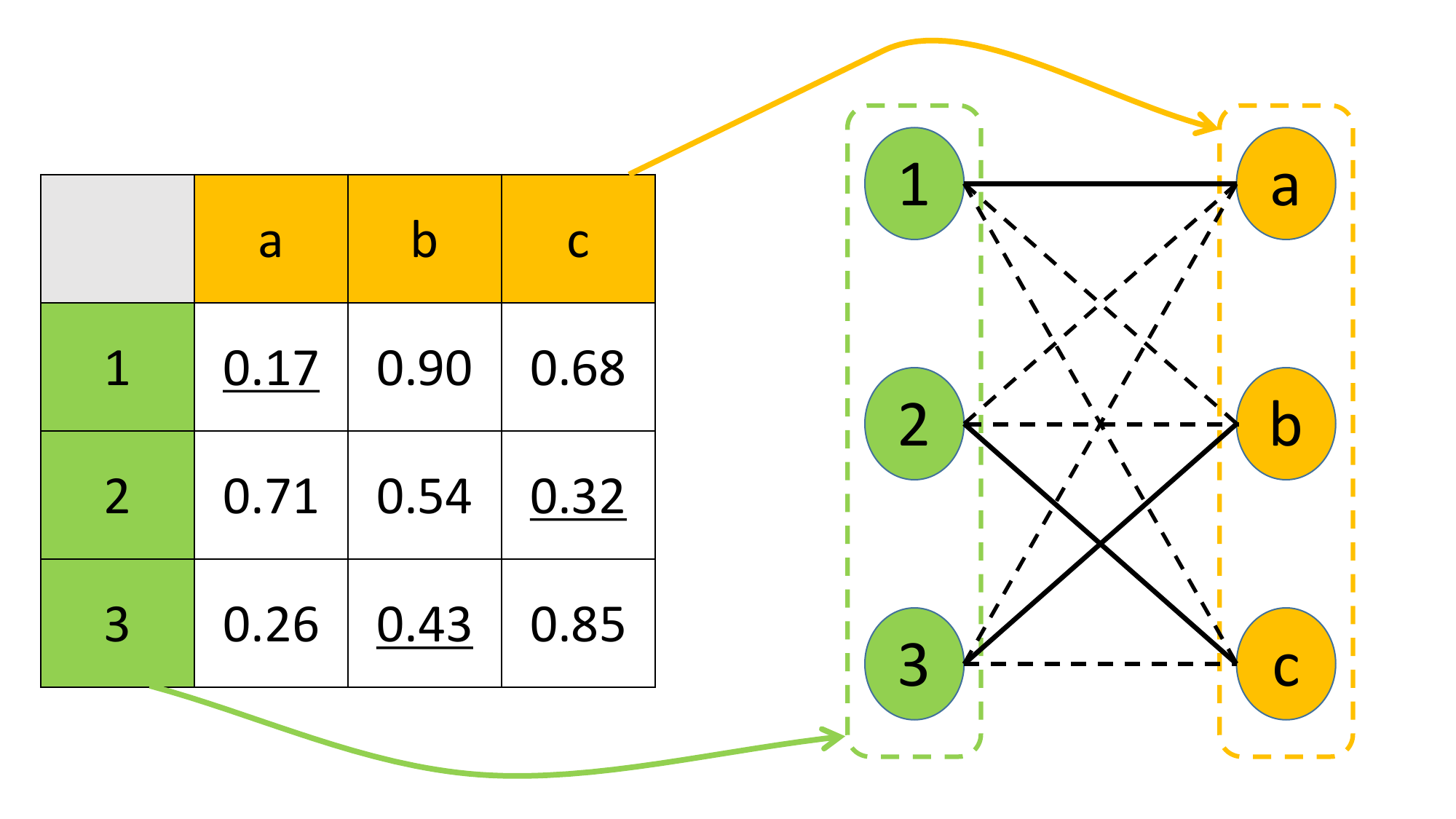}
\end{center}
\caption{Illustration of the transformation from a cost matrix to its corresponding bipartite graph. The elements underlined in the cost matrix are assigned with 1 in the assignment solution. In the constructed graph, the bold lines denote the positive edges, and the dash lines negative ones.}
\label{matrix-to-graph}
\end{figure}

Fig.~\ref{matrix-to-graph} illustrates an example of the process that transforms a cost matrix to its corresponding bipartite graph. Specifically, given a cost matrix $C$ of size $3\times{3}$, the indices colored in green and orange represent the IDs of agents and jobs respectively. Then all the agents and jobs are considered as nodes in the bipartite graph $\mathbb{G}$ and the edges are assigned with cost values according to the cost matrix. Thus finding the optimal assignment is converted to selecting reliable edges in the constructed graph $\mathbb{G}$. As shown in Fig.~\ref{matrix-to-graph}, selecting elements $C_{1a}$, $C_{2c}$, $C_{3b}$ in the cost matrix is equivalent to selecting edges $(1,a)$, $(2,c)$, $(3,b)$ from the bipartite graph.

To learn how to select reliable edges from the constructed graph $\mathbb{G}$, we build a fully trainable graph network which takes the $\mathbb{G}$ as input, and performs information aggregation and updating over the graph to predict the edge labels. The predicted label on the edge $e_{ij}$ is equivalent to the value of $x_{ij}$ in the permutation matrix which indicates whether to assign agent $i$ to job $j$ and job $j$ to agent $i$ also. The pipeline of the proposed framework is described in Sec.~\ref{section:pipeline}.

\section{Proposed Framework}\label{section:pipeline}
\begin{figure*}
\begin{center}
\includegraphics[width=1.0\linewidth]{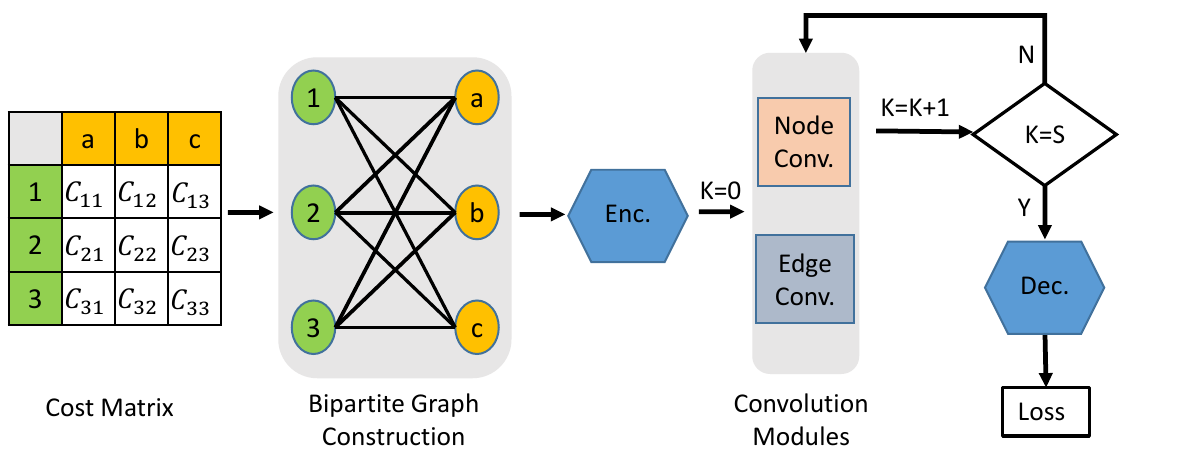}
\end{center}
\caption{The pipeline of the proposed linear assignment framework. Given a cost matrix, we first construct its corresponding bipartite graph, as that is described in Sec.~\ref{subsection:edge-labeling}. Then the constructed graph is taken as input to the encoder module. The encoder transforms the attributes on the graph to latent features. Subsequently, the high-dimension graph iteratively passes the convolution module. In this module, the edge and node convolution layers perform information aggregation and updating to calculate the structure representations for each node and edge. After $S$ iterations in the convolution module, the convolved edge attributes are passed to a decoder to predict the labels on the edges.}
\label{pipeline}
\end{figure*}

 Recently, there has been growing interest in \emph{Graph neural networks} (GNNs)~\cite{GNN_survey,GNBlock} which can well capture relational structured representations on data and has been widely applied to computer vision tasks such as action recognition~\cite{GCN_zeroshot,SKELETON} and hierarchical representation learning~\cite{Hierarchical}. Inspired by these studies, we design our learning framework by extending the \emph{graph network block}~\cite{GNBlock,WangLLJHL20} module in which some operations over graph data have been defined.
As illustrated in Fig~\ref{pipeline}, the pipeline for training the proposed framework consists of four modules: bipartite graph generation, encoder/decoder, convolution module, and loss function. Given a cost matrix, the proposed framework firstly constructs a bipartite graph that is taken as input to an encoder and transformed into a latent representation. Then the edge/node convolution layer in the convolution module iteratively performs feature aggregation from neighbors by attention/weight strategies and update the attributes for per-edge and per-node. Subsequently, the edge labels are predicted by a decoder according to the updated graph states. Finally, the loss between optimal solutions and predicted labels guides the training process. Besides, the one-to-one matching constraints are also taken into account as a part of supervision signals. In the following, we describe each module in detail.

 \subsection{Bipartite Graph Construction}
As described in Sec.~\ref{subsection:edge-labeling}, a cost matrix can be transformed into a bipartite graph in which the nodes are grouped into two classes representing agents and jobs respectively. In the constructed graph, edges only exist between the nodes in inter-class. The raw features of these edges are assigned with the cost values between source nodes and receive nodes according to the cost matrix. Therefore the initial attributes of the edges in the constructed graph can be expressed as:
 \begin{equation}
 \textbf{e}_{ij}=C_{ij}.
\label{graph_attr}
\end{equation}
Note that all the raw attributes of the nodes in the constructed graph are initialized with zero-valued vectors. In order to avoid the influence of noise edges on training, for each agent node, we keep $t$ adjacent edges with the lowest costs and remove the rest edges. The constructed graph is represented as
\begin{equation}
\mathbb{G}=(\mathbb{V},\mathbb{E},\mathcal{V},\mathcal{E}),
\end{equation}
where $\mathbb{V}$ and $\mathbb{E}$ denote the node set and edge set, $\mathcal{V}$ and $\mathcal{E}$ are the node attribute set and edge attribute set, respectively.

 \subsection{Encoder/Decoder}
 The encoder module transforms the attributes of the constructed graph into latent representations by applying an \emph{Multi-Layer Perceptrons} (MLPs) to each edge to form the embedding features. The transformed graph then is passed to the convolutional module as input to update its state.

The decoder coupled with the encoder reads out the edge attributes from the output graph and predicts each edge label through an update function.
Similarly, the update function is designed as an MLPs and mapped to each edge to form edge labels through a sigmoid activation.
 \subsection{The Convolution Module}
 The convolutional module consists of a node convolution layer and an edge convolution layer. For $i^{th}$ node in the graph, the node convolution layer collects the information from adjacent edges and its $1^{th}$ order neighborhoods by adaptive aggregation weights and updates its attributes. For each edge, the edge convolutional layer aggregates the attribute from its source node and receive node through channel-attention strategy and performs the attribute update. Although the reception field of the convolution module is $1^{th}$-order neighborhoods, the messages on each node can reach all other nodes after two iterations of convolution. Therefore, the reception field of the $k$-iteration convolution module can cover the whole graph when $k\geq2$.

 \textbf{The edge convolution layer:} This layer consists of an aggregation function and an update function. For each edge, the aggregation function collects information from its source node and receive node:
 \begin{equation}
 \bar{\mathbf{e}}_{ij}=[\mathbf{v}_i\odot{c^{v}},\mathbf{v}_j\odot{c^{v}},\mathbf{e}_{ij}\odot{c^{e}}],
 \label{edge_aggregation}
 \end{equation}
 where $\mathbf{e}_{ij}$ denotes the attributes of the edge connecting node $i$ and node $j$, $\mathbf{v}_i$ and $\mathbf{v}_j$ the attributes of $i^{th}$ and $j^{th}$ node. $\odot$ indicates the element-wise multiplication of two vector. The operator $[\cdot,\cdot,\cdot]$ concatenates its input vectors in channel wise.
 In addition, $c^{v}$ is the node channel attention vector with the same dimension as node attribute, and computed as
 \begin{equation}
 c^{v}=\kappa^{v}([max(\mathcal{V}),min(\mathcal{V}),mean(\mathcal{V})]),
 \end{equation}
 where $max$, $min$ and $mean$ return the real number vectors including maximum values, minimum values and average values of their input attribute set along channel wise, respectively.

 Similarly, the edge channel attention vector $c^{e}$ with the same dimension as edge attribute is computed as
 \begin{equation}
 c^{e}=\kappa^{e}([max(\mathcal{E}),min(\mathcal{E}),mean(\mathcal{E})]).
 \end{equation}
  The operators $\kappa^{v}$ and $\kappa^{e}$ are both parameterized as MLP with different configurations to map their input into desired space.

 Taking the concatenated attributes as input, an update function $\rho^{e}$ designed as an MLP is applied to output the updated feature:
 \begin{equation}
 \mathbf{e}_{ij}\leftarrow\rho^{e}(\bar{\mathbf{e}}_{ij}),
 \label{edge_update}
 \end{equation}

 \textbf{The node convolution layer:} This layer is designed to collect information from adjacent edges and $1^{th}$-order neighborhoods for each node. In this layer, not only the channel attention, but the aggregation weights are also taken into consideration. Specifically, for $i^{th}$ node in graph $\mathbb{G}$, this layer first aggregates the information from the associated edges and its $1^{th}$-order adjacent nodes by adaptive weights:
 \begin{equation}
 \begin{split}
 &\bar{\mathbf{v}}_{i}=\frac{1}{|\mathbb{N}_i|}\sum_{j=1}^{|\mathbb{N}_i|}\rho_{1}^{v}([\mathbf{e}_{ij}\odot{c^{e}}, \omega_{ij} ({\mathbf{v}_{j}\odot{c^{v}}})]),\\
 &\mathbf{e}_{ij}\in \mathcal{E}_i \quad \textrm{and} \quad \mathbf{v}_{j}\in \mathcal{V}_i,
 \end{split}
 \label{node_aggregation}
 \end{equation}
 where $\rho_{1}^{v}$ denotes the function to transform its input to an embedding feature. $\mathcal{E}_i$ denotes the attribute set of all edges associated with node $\mathbf{v}_{i}$ in graph $\mathbb{G}$ and $\mathcal{V}_i$ the attribute set of $1^{th}$-order adjacent nodes to node $\mathbf{v}_{i}$. For node $\mathbf{v}_{i}$, $\omega_{ij}$ is the weight measuring the contribution of its adjacent node $\mathbf{v}_{j}$ during feature aggregation, and computed as
 \begin{equation}
 \omega_{ij}=\tau([\mathbf{v}_{i},\mathbf{v}_{j}]).
 \end{equation}

 Then the collected embedding features are concatenated with the current attributes of node $\mathbf{v}_{i}$ and are passed to another transformation function that outputs the updated attributes for node $\mathbf{v}_{i}$:
 \begin{equation}
 \mathbf{v}_{i}\leftarrow\rho_{2}^{v}([\bar{\mathbf{v}}_{i},\mathbf{v}_{i}]).
 \label{node_update}
 \end{equation}

 The functions $\rho_{1}^{v}$, $\rho_{2}^{v}$ and $\tau$ are all specified as MLPs modules, while the structures and parameters in these MLPs are different from each other.

It is worth mentioning that the operations in the above two layers are independent of the size and structure of the constructed graph, which allows the proposed learning framework to be trained using cost matrices of different sizes.

 \subsection{Loss Function}
Similar to~\cite{deepmot}, we consider the linear assignment problem as a binary classification task and divide the elements in the ground-truth assignment matrix $Y^{gt}$ into positive labels and negative ones. Considering the case that for each node, there is at most one positive edge among its adjacent edges and the rest are negative ones, to avoid the negative labels dominating the training, we utilize the \emph{Balanced Cross Entropy} as the loss function:
 \begin{equation}
 \begin{aligned}
 \mathcal{L}_{A}&=-\sum_{i=1}^{n\times{t}}[w\times\mathbf{y}^{gt}_{i}\log(\mathbf{y}_{i})\\
 &+(1-w)\times(1-\mathbf{y}^{gt}_{i})\log(1-\mathbf{y}_{i})],
 \end{aligned}
 \label{binary_loss}
 \end{equation}
 where $\mathbf{y}_i$ is the predicted label for edge $(s_i,r_i)$ which connects agent $s_i$ and job $r_i$, $\mathbf{y}^{gt}$ is the corresponding ground-truth vector indicating the edges as positive and negative, and $w$ is the weight which balances the loss to avoid the negative labels dominating the training.

 The one-to-one matching constraint embodies the nature of the solution for linear assignment problems. However, it has not been fully explored in previous learning-based approaches.
 To impose this constraint in our learning framework, we first construct a predicted assignment matrix $Y\in\mathbf{R}^{n\times{n}}$ whose size is the same as the problem:
 \begin{equation}
 Y_{j,k}=\left\{
 \begin{aligned}
 &\textbf{y}_{ind(j,k)} & \textrm{if~} \exists (j,k)\in \mathbb{E},\\
 &0 & \textrm{otherwize},
 \end{aligned}
 \right.
 \end{equation}
 where $ind(\cdot , \cdot)$ is a bijection mapping an edge to an integer index.
 Then we design the soft constraint loss as follows:
 \begin{equation}
 \mathcal{L}_{1}=\|\mathbf{1}-sum_r(Y)\|_2+\|\mathbf{1}-sum_r(Y^T)\|_2,
 \label{sum_cons}
 \end{equation}
 \begin{equation}
 \mathcal{L}_{2}=\|\mathbf{1}-norm_r(Y)\|_2+\|\mathbf{1}-norm_r(Y^T)\|_2,
 \label{sparse_cons}
 \end{equation}
 \begin{equation}
 \mathcal{L}_{C}=\mathcal{L}_{1}+\mathcal{L}_{2},
 \label{cons_loss}
 \end{equation}
 In the above equations, $\mathbf{1}$ is a one-valued vector; $sum_r(\cdot)$ sums the values in predicted assignment matrix along the row-wise; $norm_r(\cdot)$ returns a vector in which each element is the 2-norm of corresponding row-vector.
 In training, Eq.~\ref{sum_cons} drives the prediction to satisfy the constraints in Eq.~\ref{constraint_1} and~\ref{constraint_2}. As a part of supervision signals in training, Eq.~\ref{sparse_cons} guides the proposed network to output sparse predictions.

 We finally combine the binary classification loss and the constraint loss to guide the training process as follows:
 \begin{equation}
 \mathcal{L}=\mathcal{L}_{A}+\alpha\mathcal{L}_{C},
 \label{loss}
 \end{equation}
 where $\alpha>0$ weights the degree of the soft one-to-one matching constraints imposed.
\section{Experiments}
\begin{table*}[!t]
\centering
\caption{Average precision (\%) on synthetic datasets $SynData^1_{10\_150}$.}
\label{table:prec_10_150}
\scalebox{0.865}{
\begin{tabular}
{@{\hspace{.3mm}}l@{\hspace{.3mm}} | @{\hspace{.7mm}}c@{\hspace{.7mm}} @{\hspace{.7mm}}c@{\hspace{.7mm}} @{\hspace{.7mm}}c@{\hspace{.7mm}} @{\hspace{.7mm}}c@{\hspace{.7mm}} @{\hspace{.7mm}}c@{\hspace{.7mm}} @{\hspace{.7mm}}c@{\hspace{.7mm}} @{\hspace{.7mm}}c@{\hspace{.7mm}} @{\hspace{.7mm}}c@{\hspace{.7mm}} @{\hspace{.7mm}}c@{\hspace{.7mm}} @{\hspace{.7mm}}c@{\hspace{.7mm}} @{\hspace{.7mm}}c@{\hspace{.7mm}} @{\hspace{.7mm}}c@{\hspace{.7mm}} @{\hspace{.7mm}}c@{\hspace{.7mm}} @{\hspace{.7mm}}c@{\hspace{.7mm}} @{\hspace{.7mm}}c@{\hspace{.7mm}}
| @{\hspace{.7mm}}c@{\hspace{.7mm}}}
  \hline
    Algorithm & 10 & 20 & 30 & 40 & 50 & 60 & 70 & 80 & 90 & 100 & 110 & 120 & 130 & 140 & 150&AVG\\
  \hline
  SH~\cite{sinkhorn} &  62.4 & 51.2 & 47.1 & 43.1 & 39.7 & 38.5 & 36.6 & 35.0 & 34.2 & 33.7 & 32.5 & 31.8 & 31.0 & 30.6 & 29.8 & 38.5 \\
  SD~\cite{sinkhorn_dis} &  69.6 &  59.9 &  56.3 &  53.1 &  50.6 &  48.8 &  46.8 &  45.5 &  44.9 &  44.1 &  43.0 &  42.5 &  41.5 &  40.9 &  40.4 &  44.6 \\
  RM~\cite{DMMNet}&  77.9 &	67.6 &	61.0 &	57.0 &	53.7 &	51.4 &	50.7 &	50.6 &	50.0 &	49.7 &	50.9 &	50.2 &	50.4 &	49.5 &	49.8 &	54.7 \\
  \hline
  DFC~\cite{Deep_lap} & 59.4 & 49.9 & 47.7 & 46.8 & 46.3 & 45.0 & 43.8 & 41.8 & 40.8 & 38.9 & 37.5 & 36.0 & 34.5 & 33.3 & 32.0 & 42.2 \\
  DCNN~\cite{Deep_lap}& 62.8 &	59.1 &	57.7 &	57.8 &	57.7 &	57.3 &	57.4 &	57.3 &	57.5 &	57.0 &	57.2 & 56.9 &	57.3 &	57.1 & 56.5 &57.8  \\
  BDL~\cite{RNN_FC}   & 63.4 &	59.8 &	60.0 &	59.1 &	59.2 &	59.0 &	58.5 &	58.9 &	58.9 &	58.4 &	58.5 &	58.2 &	58.5 &	58.3 &	57.9 &	59.1 \\
  DHN~\cite{deepmot}  & 72.3 &	67.9 &	67.0 &	65.9 &	64.9 &	64.8 &	64.4 &	63.8 &	63.2 &	62.9 &	62.8 &	62.0 &	61.8 &	60.8 &	59.7 &	64.3 \\
  \hline
  GLAN                & \textbf{79.8} &	\textbf{74.9} &	\textbf{74.4} &	\textbf{74.2} &	\textbf{73.7} &	\textbf{74.1} & \textbf{73.8} &	\textbf{73.9} &	\textbf{74.2} &	\textbf{74.0} &	\textbf{73.9} &	\textbf{73.4} &	\textbf{73.8} &	\textbf{73.7} &	\textbf{73.3} &	\textbf{74.4} \\
\hline
\end{tabular}
}
\end{table*}
\begin{table*}[!t]
\centering
\caption{Average running time (ms) on synthetic datasets $SynData^1_{10\_150}$.}
\label{table:times}
\scalebox{0.93}{
\begin{tabular}
{@{\hspace{.3mm}}l@{\hspace{.3mm}} | @{\hspace{.3mm}}c@{\hspace{.5mm}} @{\hspace{.5mm}}c@{\hspace{.5mm}} @{\hspace{.5mm}}c@{\hspace{.5mm}} @{\hspace{.5mm}}c@{\hspace{.5mm}} @{\hspace{.5mm}}c@{\hspace{.5mm}} @{\hspace{.5mm}}c@{\hspace{.5mm}} @{\hspace{.5mm}}c@{\hspace{.5mm}} @{\hspace{.5mm}}c@{\hspace{.5mm}} @{\hspace{.5mm}}c@{\hspace{.5mm}} @{\hspace{.5mm}}c@{\hspace{.5mm}} @{\hspace{.5mm}}c@{\hspace{.5mm}} @{\hspace{.5mm}}c@{\hspace{.5mm}} @{\hspace{.5mm}}c@{\hspace{.5mm}} @{\hspace{.5mm}}c@{\hspace{.5mm}} @{\hspace{.5mm}}c@{\hspace{.5mm}}}
  \hline
    Algorithm & 10 & 20 & 30 & 40 & 50 & 60 & 70 & 80 & 90 & 100 & 110 & 120 & 130 & 140 & 150\\
  \hline
  SH~\cite{sinkhorn} &  0.02 &0.06& 0.05 & 0.02 & 0.02 & 0.03 & 0.03& 0.04 &0.05 & 0.04 &0.05 & 0.06 &0.05& 0.07 & 0.07\\
  SD~\cite{sinkhorn_dis} &  0.01 &0.01& 0.01 & 0.01 & 0.01 & 0.01 & 0.01& 0.01 &0.01 & 0.01 & 0.01 & 0.01 &0.01 & 0.01 & 0.01 \\
  RM~\cite{DMMNet}& 1.6k & 1.7k & 1.7k& 1.8k & 1.8k & 2.0k& 2.2k& 2.4k& 2.6k & 2.7k& 3.1k & 3.6k & 4.3k &4.8k & 5.6k\\
  \hline
  DFC~\cite{Deep_lap}  & 0.54&0.55&0.58&0.57&0.56&0.59&0.60&0.60&0.60&0.57&0.56&0.57&0.56&0.55&0.62 \\
  DCNN~\cite{Deep_lap} & 0.36&0.40&0.48&0.46&0.42&0.39&0.43&0.35&0.49&0.49&0.47&0.45&0.56&0.53&0.57 \\
  BDL~\cite{RNN_FC}    & 3.21&4.41&5.85&7.19&9.78&11.6&12.8&14.2&15.6&13.6&14.8&16.0&17.3&18.4&19.7\\
  DHN~\cite{deepmot}   & 196&308& 520 & 842 & 1.4k& 1.9k& 2.5k& 3.1k & 4.0k & 5.1k& 6.8k &7.3k & 9.4k& 10.8k & 11.3k \\
  \hline
  GLAN&  6& 7 & 7 & 7 &7 &7 & 7&7 & 7 & 7 & 7& 7 &7& 8 &8 \\
\hline
\end{tabular}
}
\end{table*}

\subsection{Settings}
In our experiments, we set the maximum of iteration in the convolution module and the dimension of edge feature to 5 and 16, respectively. And the number of adjacent edges for each node $t$ is assigned with 8. In addition, the weight for positive edges is fixed to 0.9.
\subsection{Training}\label{sec:training}
For training the proposed framework, we generate a synthetic dataset that consists of $M$ samples. Each sample is composed of a cost matrix $C$ in which the elements are generated from a uniform distribution on $(0,1)$ and the corresponding optimal assignment solution $Y^{gt}$ which is obtained by the Hungarian algorithm~\cite{hungarian}. In experiments, the cost matrices in the synthetic dataset have varying sizes ranging from 10 to 150 with an interval of 10. For the sake of clarity, with the default lower bound of cost data as 0, we name the synthetic dataset $SynData^1_{10\_150}$ where the superscript 1 denotes the upper bound of the cost value and the subscript 10\_150 represents the minimum size (10) and maximum size (150) of cost matrices. Besides, we randomly sample 30\% of this dataset for evaluation and keep the other 70\% for training. The training totally takes 20 epochs where the learning rate is set as 0.003 initially and declined by 5\% after each 5 epochs. In addition, the $\alpha$ in Eq.~\ref{loss} is 0 at the beginning and increases 0.01 after each epoch.

\subsection{Evaluation on Synthetic Datasets}

\subsubsection{Comparison with Baseline Methods}

In this section, we report the experimental results of the proposed framework, named GLAN (\emph{Graph-based Linear Assignment Network}), compared with six state-of-the-art baselines, SH~\cite{sinkhorn}, SD~\cite{sinkhorn_dis}, RM~\cite{DMMNet}, DLASP~\cite{Deep_lap}, BDL~\cite{RNN_FC}, and DHN~\cite{deepmot} of which the first three algorithms are traditional learning-free methods and the last three are deep learning approaches. In Tables~\ref{table:prec_10_150} and~\ref{table:times}, two versions of DLASP~\cite{Deep_lap} are involved in comparison, named DFC and DCNN which are build upon \emph{Fully-Connection} (FC) layers and CNNs~\cite{CNN} respectively.

To evaluate the performance of these methods, we apply \emph{Greedy discretization} to the solution matrices predicted by these approaches, and define the assignment precision criteria as
\begin{equation}
p=\frac{tr(Y^\mathrm{T}Y^{gt})}{tr((Y^{gt})^\mathrm{T}Y^{gt})},
\end{equation}
where $tr(\cdot)$ returns the trace of given matrix. In addition, in order to fairly compare our method with baselines in terms of computation efficiency, all the methods are run on PC with a Intel Core i5-4590 CPU (3.30 GHz), a RAM (16GB) and a GTX 1060Ti GPU (6G).

The average precision of each tested algorithm is illustrated in Table~\ref{table:prec_10_150}. In addition, Table~\ref{table:times} reports the comparisons in terms of computational time. In non-learnable approaches, adopting simple normalization along row and column, SinkHorn~\cite{sinkhorn} can not obtain satisfactory assignment precision, despite of its relative fast inference process. As a variant of SinkHorn~\cite{sinkhorn} with entropic penalization, SD~\cite{sinkhorn_dis} runs with the lowest time cost, and significantly outperforms SinkHorn~\cite{sinkhorn} in assignment precision. However, their performance on assignment precision often deteriorates when the size of problems increases. The RM~\cite{DMMNet} algorithm employs an elaborate reasoning process to update the solution iteratively and achieves the best assignment precision among the learning-free baselines. However, its computation time grows greatly with the increase of the problem size.
In learning-based methods, DFC~\cite{Deep_lap} passes the sub-classification tasks generated from the cost matrix to several FC layers to make decisions for linear assignment problems. Its computational complexity is the lowest than other learning-based methods, but it gets the worst assignment precision. Similar to the DFC~\cite{Deep_lap}, decomposing the linear assignment problem into a sequence of smaller sub-assignment tasks, the RNN-based model BDL~\cite{RNN_FC} achieves a better performance on assignment precision due to its adaptation to problems of varying sizes. However, the relationships between sub-assignment tasks are not taken into account in both of their frameworks. Employing several stacked CNN layers, DCNN~\cite{Deep_lap} also achieves a better performance than the FC-based version of DLASP~\cite{Deep_lap}, i.e., DFC. However, due to the limited receptive field of convolution kennels, it can not makes assignment decisions through global information when the size of the problem is greater than its receptive field. To receive information from the global scope, another RNN-based model, named DHN~\cite{deepmot}, transforms the cost matrix into two sequential patterns from column-wise and row-wise respectively to make assignment decisions. The performance of DHN~\cite{deepmot} outperforms all the other previous baselines on assignment precision. But same as the RM~\cite{DMMNet}, the computational time of DHN~\cite{deepmot} is also sensitive to the problem size.

Devoting to learning the inductive representations and relational structures, that help to make the decisions for linear assignment problems, our graph network achieves the best assignment precisions in general as shown in Table~\ref{table:prec_10_150}. Specifically, the average assignment precision of the proposed method superiors the one of DHN~\cite{deepmot} by $10.08\%$.
By multiple iterations of the convolution module, the receptive field of our model can cover the whole graph no matter how large the graph is.
As shown Table~\ref{table:prec_10_150}, our method achieves consistently high assignment accuracies with the problem size increasing.
Furthermore, since the network architecture is independent of the structure and size of the graph, the running time of the proposed method increases very slightly as the problem size expands.

\begin{figure*}[t]
\centering
  \subfigure[varying layers.]{
    \begin{minipage}[t]{0.45\linewidth}
    \centering
    \label{subfig:layers}
    \includegraphics[width=6cm,height=4cm]{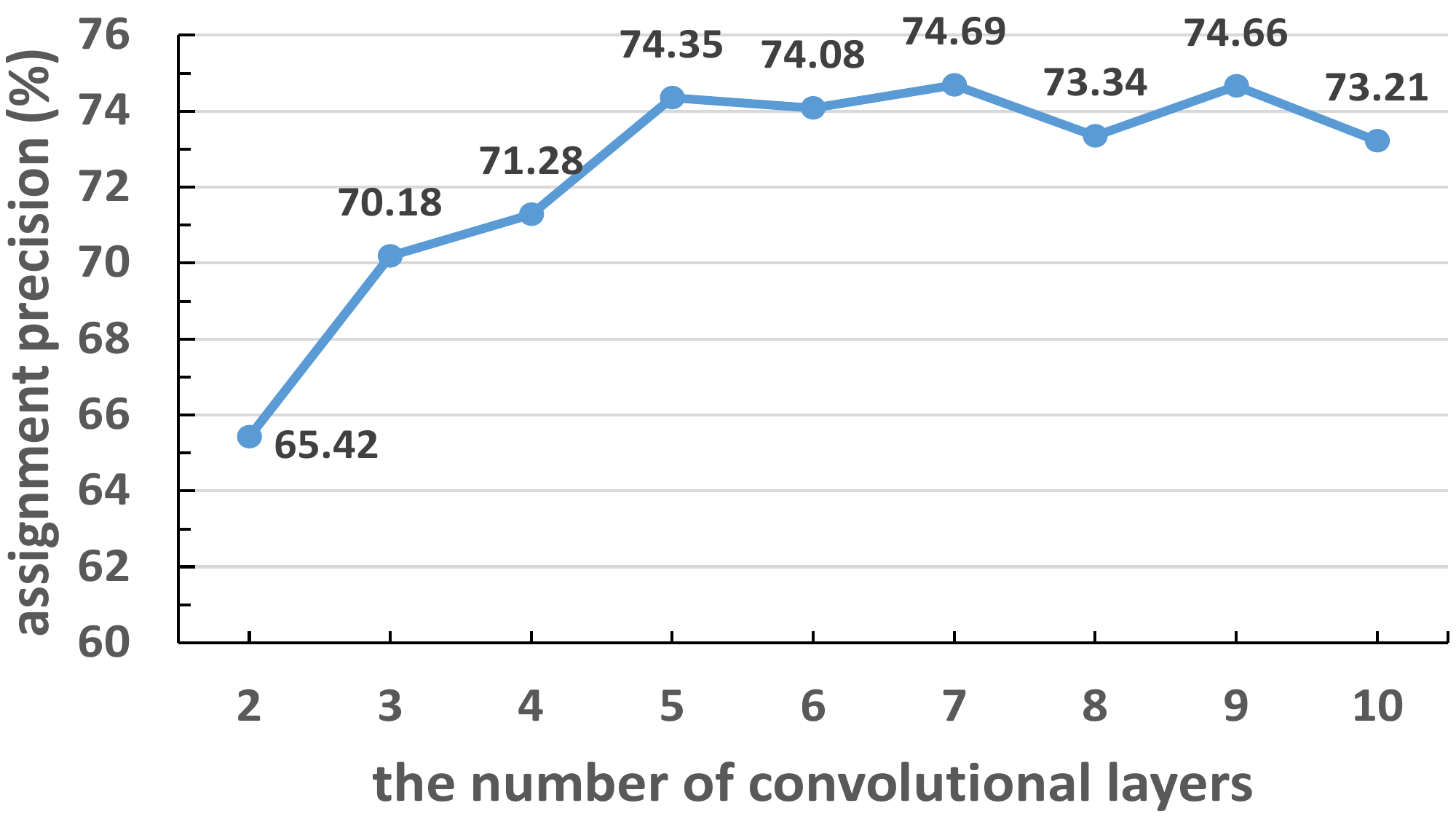}
    \end{minipage}
    }
  \subfigure[varying dimension.]{
    \begin{minipage}[t]{0.45\linewidth}
    \centering
    \label{subfig:dims}
    \includegraphics[width=6cm,height=4cm]{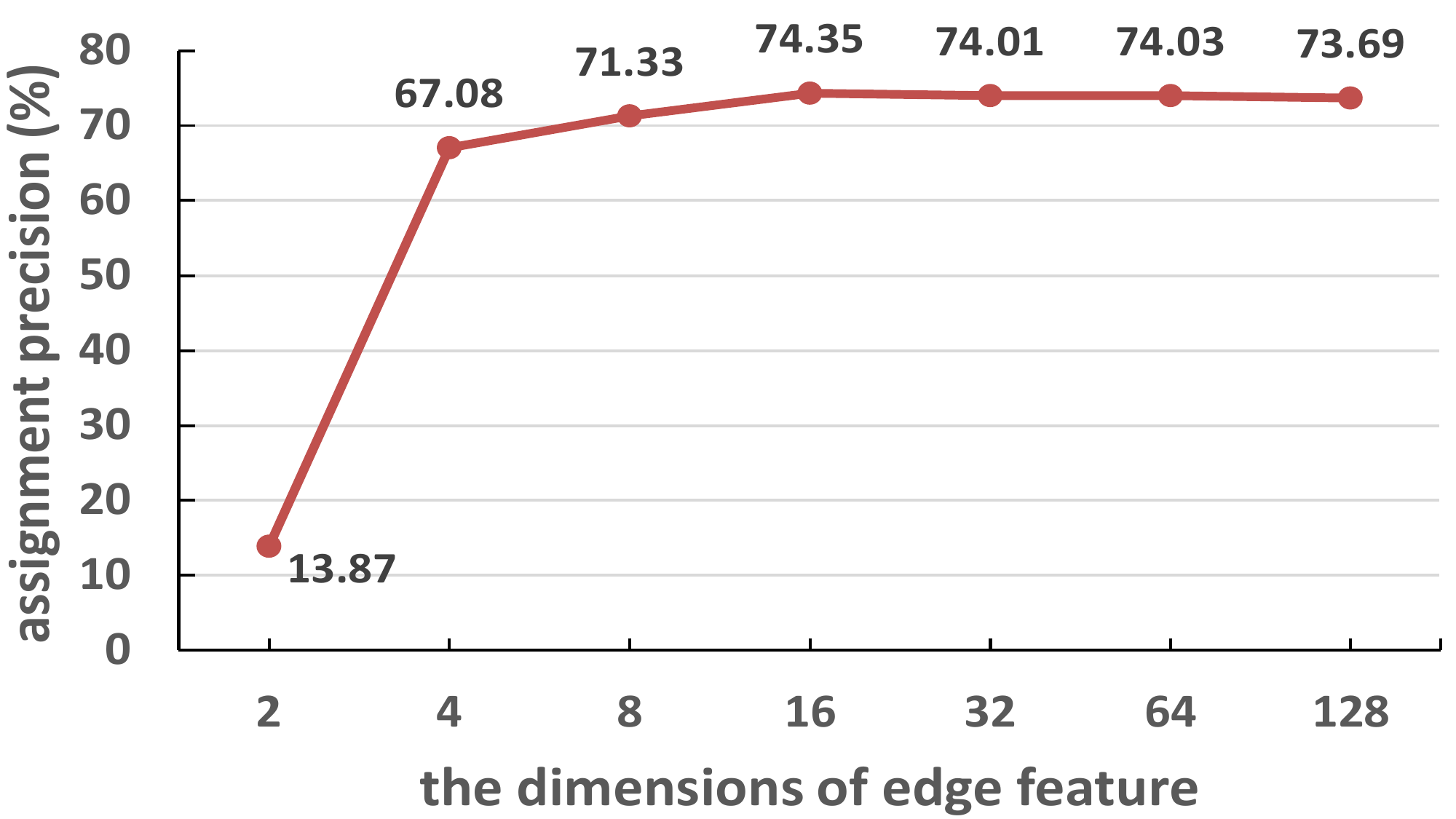}
    \end{minipage}}

  \subfigure[channel attention and aggregation weights.]{
    \begin{minipage}[t]{0.45\linewidth}
    \centering
    \label{subfig:weight}
    \includegraphics[width=6cm,height=4cm]{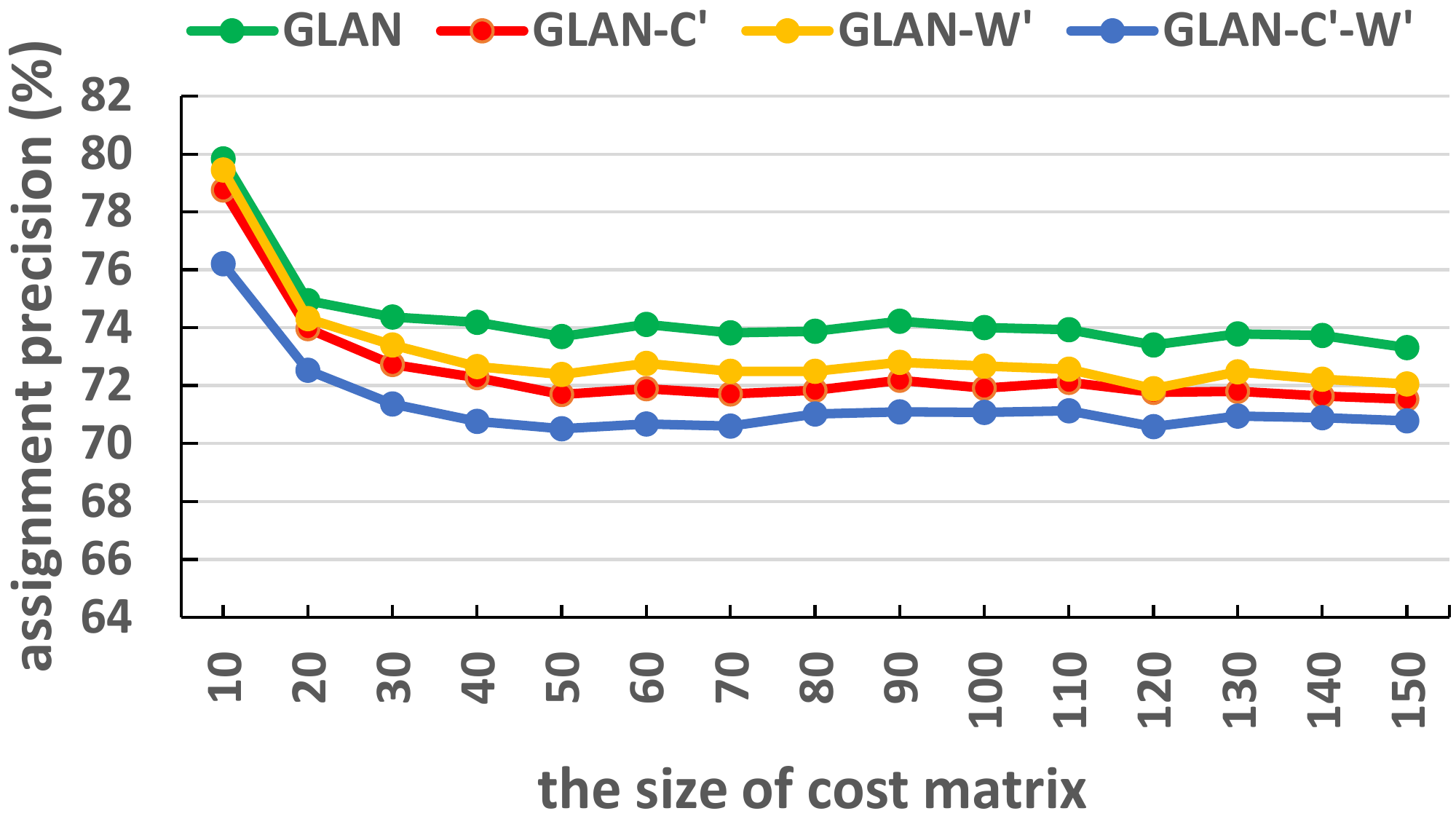}
    \end{minipage}}
  \subfigure[different constraints.]{
    \begin{minipage}[t]{0.45\linewidth}
    \centering
    \label{subfig:loss}
    \includegraphics[width=6cm,height=4cm]{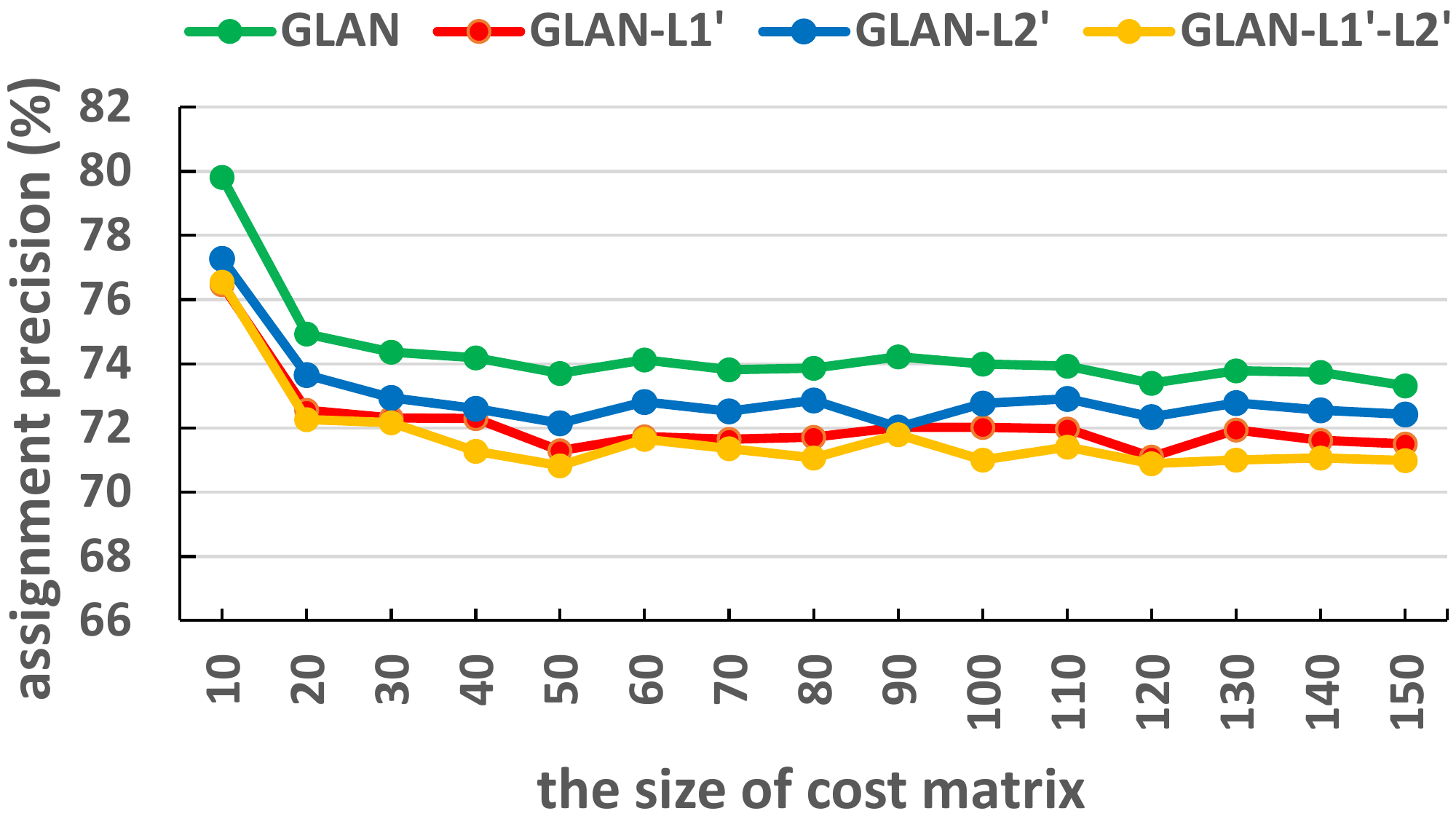}
    \end{minipage}}
  \caption{Illustration of results in ablation experiments for (a) the number of convolution layers, (b) the dimension of edge feature, (c) the channel attention and aggregation weights and (d) the constraint losses.}
  \label{fig:examples}
\end{figure*}

\subsubsection{Ablation Study}
To verify the effectiveness of each component in our framework, we perform several ablation experiments for the number of convolution layers, the dimension of edge feature, the channel attention and aggregation weights, and the constraint losses.

$\bullet$ \textbf{Influence of the number of convolution layers.}
The number of convolution layers determines the iterations of the message passing in the assignment graph. As mentioned in Sec~\ref{sec:intor}, the information of one node can be passed to any other nodes after two convolution operations. To explore the influence of the number of convolution layers, we set the dimension of edge features to 16 and vary the convolution iterations from 2 to 10. As illustrated in Fig.~\ref{subfig:layers}, when the iteration is set to 2, our model has outperformed the best baseline method DHN~\cite{deepmot}. As the number of convolution layer increases from 2 to 5, the performance of our model is improved significantly. However, there is no significant improvement when convolution iteration is set to a value greater than 5, which means that using 5 convolutional layers is sufficient to capture the global inductive information for making assignment decisions.

$\bullet$ \textbf{Influence of edge feature dimension.}
The edge feature here refers to the latent representations of edges in the convolution layer. We fix the convolution iterations to 5 and perform the ablation study for edge feature dimension, where the dimension is set to several values to illustrate its influence on performance. As illustrated in Fig.~\ref{subfig:dims}, the performance of our model is significantly improved when the dimension is expanded from 2 to 4. In addition, the performance is enhanced to a certain extent by setting the dimension to 16. However, when setting the dimension to a larger value than 16, such as 32, 64, and 128, the assignment precision is basically stable around the result predicted by assigning the dimension with 16. As mentioned above, our model is a lightweight framework, where only 5 convolution layers and 16-dimension latent representations are sufficient to achieve the nearly best assignment precision.

$\bullet$ \textbf{Effectiveness of channel attention and aggregation weights.}
During convolution process, the channel attention adaptively guides the proposed framework focus on the more important features in channel wise. For a node, the aggregation weights are used to measure the contribution of its adjacent nodes in node feature aggregation. In order to validate their effectiveness for learning structured representations to make assignment decision, we design three downgraded version of our model, named GLAN-C$^\prime$, GLAN-W$^\prime$ and GLAN-C$^\prime$-W$^\prime$, where C$^\prime$ and W$^\prime$ mean removing the channel attention and aggregation weights in convolution layer respectively. As illustrated in Fig.~\ref{subfig:weight}, both channel attention and aggregation weights can improve the assignment precision of GLAN-C$^\prime$-W$^\prime$ in all cases with different data sizes, which demonstrates that both of them have positive effects on making assignment decision. Furthermore, with the combination of channel attention and aggregation weights, the full version of framework GLAN surpasses all downgraded models and achieves the best performance in all cases. It demonstrates that the channel attention and aggregation weights are not conflicting with each other, and the combination of them can further improve the assignment precision.

$\bullet$ \textbf{Effectiveness of constraint losses.}
The loss function Eq.~\ref{sum_cons} leads the assignment matrix output by our framework to satisfy the constraints in Eq.~\ref{constraint_1} and Eq.~\ref{constraint_2}. And another loss function Eq.~\ref{sparse_cons} guides our framework to make a sparse assignment decision.
Aiming to quantify the contribution of the proposed soft constraint losses formulated as Eq.~\ref{sum_cons} and~\ref{sparse_cons}, we also design three downgraded versions of our model, named GLAN-L1$^\prime$-L2$^\prime$, GLAN-L1$^\prime$ and GLAN-L2$^\prime$, where L1$^\prime$ and L2$^\prime$ denote removing constraint loss Eq.~\ref{sum_cons} and Eq.~\ref{sparse_cons} in training respectively. As illustrated in Fig.~\ref{subfig:loss}, without any constraint loss as the supervision signal in the training process, the model GLAN-L1$^\prime$-L2$^\prime$ has the worst performance compared with other versions. With the constraint loss Eq.~\ref{sum_cons} or Eq.~\ref{sparse_cons} as a part of training supervision, the proposed framework make better assignment precisions than GLAN-L1$^\prime$-L2$^\prime$ in all cases. Furthermore, the two constraint losses are not conflicting with each other, and they are combined into the full version of our model GLAN making it achieve the best performance in all cases.
\begin{table*}[t]
\centering
\caption{Average precision (\%) on synthetic datasets $SynData^1_{200\_3000}$.}
\label{table:prec_200_3000}
\scalebox{0.865}{
\begin{tabular}
{@{\hspace{.3mm}}l@{\hspace{.3mm}} | @{\hspace{.7mm}}c@{\hspace{.7mm}} @{\hspace{.7mm}}c@{\hspace{.7mm}} @{\hspace{.7mm}}c@{\hspace{.7mm}} @{\hspace{.7mm}}c@{\hspace{.7mm}} @{\hspace{.7mm}}c@{\hspace{.7mm}} @{\hspace{.7mm}}c@{\hspace{.7mm}} @{\hspace{.7mm}}c@{\hspace{.7mm}} @{\hspace{.7mm}}c@{\hspace{.7mm}} @{\hspace{.7mm}}c@{\hspace{.7mm}} @{\hspace{.7mm}}c@{\hspace{.7mm}} @{\hspace{.7mm}}c@{\hspace{.7mm}} @{\hspace{.7mm}}c@{\hspace{.7mm}} @{\hspace{.7mm}}c@{\hspace{.7mm}} @{\hspace{.7mm}}c@{\hspace{.7mm}} @{\hspace{.7mm}}c@{\hspace{.7mm}}
| @{\hspace{.7mm}}c@{\hspace{.7mm}}}
  \hline
    Algorithm & 200 & 400 & 600 & 800 & 1k & 1.2k & 1.4k & 1.6k & 1.8k & 2k & 2.2k & 2.4k & 2.6k & 2.8k & 3k & AVG\\
  \hline
  SH~\cite{sinkhorn} &  27.3 &  23.3 &  20.7 &  19.2 &  18.0 &  17.2 &  16.6 &  15.8 &  15.3 &  14.9 &  14.5 &  14.0 &  13.9 &  13.6 &  13.2 &  15.3\\
  SD~\cite{sinkhorn_dis}&  37.3 &  32.2 &  29.0 &  26.9 &  25.6 &  24.3 &  23.3 &  22.4 &  21.8 &  21.0 &  20.5 &  19.8 &  19.6 &  19.2 &  18.6 &  21.5\\
  RM~\cite{DMMNet}&   51.1 &  48.0 &  45.8 &  46.1 &  45.9 &  45.4 &  47.1 &  46.2 &  43.8 &  42.1 &  42.2 &  41.3 &  40.6 &  39.3 &  38.7 &  42.5\\
  \hline
  DCNN~\cite{Deep_lap}& 56.8 &  55.2 &  53.9 &  52.7 &  51.6 &  50.5 &  49.6 &  48.7 &  47.8 &  47.0 &  46.0 &  45.0 &  44.1 &  42.7 &  41.5 &  48.9  \\
  BDL~\cite{RNN_FC}   & 57.3 &  54.7 &  53.1 &  52.0 &  50.9 &  49.9 &  48.9 &  47.8 &  46.7 &  46.0 &  44.8 &  43.7 &  42.5 &  41.4 &  39.9 & 48.0\\
  DHN~\cite{deepmot}  & 52.5 &	39.6 &	29.7 &	21.0 &	19.2 &	- &	- &	- &	- &	- &	- &	- &	- &	- &	- &	- \\
  \hline
  GLAN                & \textbf{73.8} &  \textbf{73.4} &  \textbf{73.6} &  \textbf{73.3} &  \textbf{73.5} &  \textbf{73.4} &  \textbf{73.2} &  \textbf{73.1} &  \textbf{73.0} &  \textbf{73.0} &  \textbf{72.7} &  \textbf{72.1} &  \textbf{71.3} &  \textbf{70.3} &  \textbf{68.4} &  \textbf{72.5} \\
\hline
\end{tabular}
}
\end{table*}

\begin{table*}[t]
\centering
\caption{Average precision (\%) on synthetic datasets $SynData^{10}_{10\_150}$.}
\label{table:prec_10_150_scale}
\scalebox{0.865}{
\begin{tabular}
{@{\hspace{.3mm}}l@{\hspace{.3mm}} | @{\hspace{.7mm}}c@{\hspace{.7mm}} @{\hspace{.7mm}}c@{\hspace{.7mm}} @{\hspace{.7mm}}c@{\hspace{.7mm}} @{\hspace{.7mm}}c@{\hspace{.7mm}} @{\hspace{.7mm}}c@{\hspace{.7mm}} @{\hspace{.7mm}}c@{\hspace{.7mm}} @{\hspace{.7mm}}c@{\hspace{.7mm}} @{\hspace{.7mm}}c@{\hspace{.7mm}} @{\hspace{.7mm}}c@{\hspace{.7mm}} @{\hspace{.7mm}}c@{\hspace{.7mm}} @{\hspace{.7mm}}c@{\hspace{.7mm}} @{\hspace{.7mm}}c@{\hspace{.7mm}} @{\hspace{.7mm}}c@{\hspace{.7mm}} @{\hspace{.7mm}}c@{\hspace{.7mm}} @{\hspace{.7mm}}c@{\hspace{.7mm}}
| @{\hspace{.7mm}}c@{\hspace{.7mm}}}
  \hline
    Algorithm & 10 & 20 & 30 & 40 & 50 & 60 & 70 & 80 & 90 & 100 & 110 & 120 & 130 & 140 & 150&AVG\\
  \hline
  SH~\cite{sinkhorn} &  62.4 &	51.2 &	47.1 &	43.1 &	39.7 &	38.5 &	36.6 &	35.0 &	34.2 &	33.7 &	32.5 &	31.8 &	31.0 &	30.6 &	29.8 & 38.5 \\
  SD~\cite{sinkhorn_dis} &  76.1 &  68.0 &  63.7 &  58.9 &  56.0 &  54.5 &  53.2 &  51.0 &  50.3 &  48.9 &  47.9 &  47.1 &  46.3 &  45.7 &  44.9 &  49.8 \\
  RM~\cite{DMMNet}&     \textbf{87.3} &	\textbf{86.4} &	\textbf{75.5} & 68.4 &	68.0 &	65.1 &	64.1 &	60.9 &	59.2 &	59.8 &	59.4 &	58.9 &	58.3 &	57.2 &	52.1  & 65.4
\\
  \hline
  DFC~\cite{Deep_lap} & 52.7 &	42.6 &	41.7 &	36.6 &	23.8 &	26.0 &	23.2 &	20.5 &	18.3 &	17.4 &	16.6 &	14.1 &	12.2 &	12.4 &	11.1  & 24.6
 \\
  DCNN~\cite{Deep_lap}& 49.9 &	51.9 &	52.2 &	54.5 &	55.1 &	56.2 &	56.1 &	55.5 &	55.5 &	55.4 &	56.4 &	55.0 &	56.3 &	55.9 &	55.8  & 54.8
  \\
  BDL~\cite{RNN_FC}   & 49.5 &	54.2 &	54.7 &	57.0 &	56.1 &	57.4 &	57.0 &	56.7 &	57.3 &	57.2 &	57.5 &	57.2 &	57.6 &	57.1 &	57.5  & 56.3
 \\
  DHN~\cite{deepmot}  & 54.6 &	54.6 &	53.6 &	52.1 &	49.4 &	49.3 &	47.3 &	46.5 &	45.5 &	44.1 &	43.1 &	42.4 &	42.7 &	40.0 &	39.7  & 47.0
 \\
  \hline
  GLAN                & 80.0 &	74.9 &	74.3 &	\textbf{74.1} &	\textbf{73.7} &	\textbf{74.2} &	\textbf{73.8} &	\textbf{73.9} &	\textbf{74.3} &	\textbf{74.0} &	\textbf{73.9} &	\textbf{73.4} &	\textbf{73.8} &	\textbf{73.8} &	\textbf{73.3} &	\textbf{74.4} \\
\hline
\end{tabular}
}
\end{table*}

\begin{table*}[t]
\centering
\caption{Average precision (\%) on synthetic datasets $SynData^{10}_{200\_3000}$.}
\label{table:prec_200_3000_scale}
\scalebox{0.865}{
\begin{tabular}
{@{\hspace{.3mm}}l@{\hspace{.3mm}} | @{\hspace{.7mm}}c@{\hspace{.7mm}} @{\hspace{.7mm}}c@{\hspace{.7mm}} @{\hspace{.7mm}}c@{\hspace{.7mm}} @{\hspace{.7mm}}c@{\hspace{.7mm}} @{\hspace{.7mm}}c@{\hspace{.7mm}} @{\hspace{.7mm}}c@{\hspace{.7mm}} @{\hspace{.7mm}}c@{\hspace{.7mm}} @{\hspace{.7mm}}c@{\hspace{.7mm}} @{\hspace{.7mm}}c@{\hspace{.7mm}} @{\hspace{.7mm}}c@{\hspace{.7mm}} @{\hspace{.7mm}}c@{\hspace{.7mm}} @{\hspace{.7mm}}c@{\hspace{.7mm}} @{\hspace{.7mm}}c@{\hspace{.7mm}} @{\hspace{.7mm}}c@{\hspace{.7mm}} @{\hspace{.7mm}}c@{\hspace{.7mm}}
| @{\hspace{.7mm}}c@{\hspace{.7mm}}}
  \hline
    Algorithm & 200 & 400 & 600 & 800 & 1k & 1.2k & 1.4k & 1.6k & 1.8k & 2k & 2.2k & 2.4k & 2.6k & 2.8k & 3k & AVG\\
  \hline
  SH~\cite{sinkhorn} &  27.3 &  23.3 &  20.7 &  19.2 &  18.0 &  17.2 &  16.6 &  15.8 &  15.3 &  14.9 &  14.5 &  14.0 &  13.9 &  13.6 &  13.2 &  15.3\\
  SD~\cite{sinkhorn_dis}&  42.0 &  35.8 &  32.0 &  29.7 &  28.1 &  26.5 &  25.9 &  24.5 &  23.9 &  23.1 &  22.5 &  21.9 &  21.4 &  20.7 &  20.3 &  23.6\\
  RM~\cite{DMMNet}&     57.5 &  53.3 &  47.0 &  47.1 &  46.8 &  46.4 &  50.8 &  49.0 &  45.6 &  43.5 &  43.8 &  42.7 &  41.8 &  41.4 &  40.4 &  44.3\\
  \hline
  DCNN~\cite{Deep_lap}& 54.9 &  57.3 &  54.9 &  57.0 &  56.9 &  53.7 &  54.4 &  53.6 &  52.2 &  51.6 &  51.0 &  49.8 &  47.3 &  48.4 &  46.1 &  52.6 \\
  BDL~\cite{RNN_FC}   & 56.0 &  56.1 &  54.7 &  55.6 &  55.0 &  50.3 &  50.5 &  44.0 &  41.5 &  43.7 &  40.8 &  41.0 &  42.4 &  36.5 &  37.3   & 47.0 \\
  DHN~\cite{deepmot}  & 35.0 &	26.2 &	16.7 &	18.5 &	11.9 &	- &	- &	- &	- &	- &	- &	- &	- &	- & -  & - \\
  \hline
  GLAN                & \textbf{73.9} &  \textbf{73.3} &  \textbf{73.6} &  \textbf{73.4} &  \textbf{73.4} &  \textbf{73.5} &  \textbf{73.2} &  \textbf{73.2} &  \textbf{73.0} &  \textbf{73.1} &  \textbf{72.7} &  \textbf{72.1} &  \textbf{71.3} &  \textbf{70.3} &  \textbf{68.4} &  \textbf{72.6} \\
\hline
\end{tabular}
}
\end{table*}

\subsection{generalization study}
Generalization ability is an important quality of linear assignment solvers, especially for learning-based methods that are usually sensitive to the different data distribution between test data and training data. To verify the generalization ability of our framework and baselines, we train the learning-based models on the synthetic dataset $SynData^1_{10\_150}$ and test them on three types of dataset which have different data distributions from $SynData^1_{10\_150}$.

Specifically, in order to evaluate the assignment precision on dataset with larger cost matrices, we generate a synthetic dataset named $SynData^1_{200\_3000}$ where the problem size is ranging from 200 to 3000 with an interval of 200 and the cost value is in (0,1). Table~\ref{table:prec_200_3000} reports the assignment precisions of our framework compared with the state-of-the-art baselines. In traditional solvers, the performance of SH~\cite{sinkhorn} and SD~\cite{sinkhorn_dis} have an evident tendency to degradation with the problem size expanding. The assignment precision of the other traditional method, i.e., RM~\cite{DMMNet}, is not sensitive to the problem size, however its performance is still poor.
In the learning-based method, with the expansion of the problem scale, the assignment precision of DHN is severely reduced. When the problem size is greater than 1000, the problem cannot be solved within 5 minutes, which makes it difficult to perform complete statistics on the results of large-sized problems.
In addition, DCNN~\cite{Deep_lap} and BDL~\cite{RNN_FC} have similar average assignment precision, while their performance are still sensitive to the size of problem. Different from these baselines discussed above, our proposed framework GLAN achieves consistent assignment precision, and surpasses all baselines on all cases, which demonstrates that our model trained on small-sized data is competent for more difficult problems with large-sized cost.

Besides, we also test all the methods on the dataset which is the combination of the datasets $SynData^{1}_{10\_150}$ and $SynData^{1}_{200\_3000}$, yet the cost matrices are multiplied by a real value randomly sampled from 1 to 10 making the cost value be in $(0,10)$.
For the sake of clarity, this dataset is divided into two parts named $SynData^{10}_{10\_150}$ and $SynData^{10}_{200\_3000}$, and the corresponding experimental results are reported in Table~\ref{table:prec_10_150_scale} and Table~\ref{table:prec_200_3000_scale}. Concretely, Table~\ref{table:prec_10_150_scale} illustrates the comparison results to validate the models' generalization on test data which has a different cost value interval from the training data. And Table~\ref{table:prec_200_3000_scale} reports the experimental results predicted on a more difficult dataset in which not only the problem size but the cost value interval are also different from the training data. In traditional solvers, SH~\cite{sinkhorn} simply makes assignment decision by performing column- and row- normalization iteratively, thus is not sensitive to the cost value scale and achieves the similar assignment precision to its results on $SynData^{1}_{10\_150}$ and $SynData^{1}_{200\_3000}$. As the variant of SH~\cite{sinkhorn}, SD~\cite{sinkhorn_dis} achieves better performance on the dataset with interval of $(0,10)$ than that in $(0,1)$. Interestingly, RM~\cite{DMMNet} achieves the best assignment precision on data with varying sizes from 10 to 30 in Table~\ref{table:prec_10_150_scale}, because multiplying a value greater than 1 on cost matrix is equivalent to expanding the gradient during optimization, thus speeding up its convergence rate. Besides, in the case of the same data size, RM~\cite{DMMNet} achieves better results on data in $(0,10)$ than on data in $(0,1)$. However, the assignment precision of RM~\cite{DMMNet} is lower than our proposed framework GLAN when the problem size is greater than 30. In learning-based baselines, both DCNN~\cite{Deep_lap} and BDL~\cite{RNN_FC} obtain relative low assignment precision on data with small-size such as 10 and 20, and gradually increase the precision until it is stable around a certain value as the problem size expands. The results of DCNN~\cite{Deep_lap} and BDL~\cite{RNN_FC} on $SynData^{10}_{10\_150}$ demonstrate that both of them overfit small-sized training data and are susceptible to different data distribution. Besides, the assignment precision of DFC~\cite{Deep_lap} and DHN~\cite{deepmot} on $SynData^{10}_{10\_150}$ are obviously worse than the results on $SynData^{1}_{10\_150}$. In addition, the performance of DHN~\cite{deepmot} on $SynData^{10}_{200\_3000}$ are still deteriorated as the problem size increases, and worse than that on $SynData^{1}_{160\_300}$, which demonstrate that DHN~\cite{deepmot} fails to handle the linear assignment problem whose data size and value scale are different from its training data. Due to the structured inductive representations, our framework GLAN obtains the similar performance on $SynData^{10}_{10\_150}$ as that on the dataset $SynData^{1}_{10\_150}$, and achieves the best average assignment precision. Furthermore, it also obtains the best and nearly consistent assignment precision on $SynData^{10}_{200\_3000}$ where both the problem size and cost value scale are different from the training data.

The experimental results mentioned above demonstrate that our framework is insensitive to the data size and value scale, and can achieve consistent assignment precision.

\subsection{Evaluation in MOT.}

\begin{table*}[!t]
\centering
\caption{Tracking performance on MOTChallenge benchmarks. The arrows indicate low or high optimal metric values.}
\label{table:MOT_all}
\begin{spacing}{1.1}
\begin{tabular}
{ @{\hspace{.7mm}}c@{\hspace{.7mm}}@{\hspace{7.mm}}l@{\hspace{.7mm}}  @{\hspace{.7mm}}c@{\hspace{.7mm}} @{\hspace{.7mm}}c@{\hspace{.7mm}} @{\hspace{.7mm}}c@{\hspace{.7mm}} @{\hspace{.7mm}}c@{\hspace{.7mm}} @{\hspace{.7mm}}c@{\hspace{.7mm}} @{\hspace{.7mm}}c@{\hspace{.7mm}} @{\hspace{.7mm}}c@{\hspace{.7mm}} @{\hspace{.7mm}}c@{\hspace{.7mm}} }
  \hline
  &Method & MOTA$\uparrow$ & MOTP$\uparrow$ & IDF1$\uparrow$ & MT$\uparrow$ & ML$\downarrow$&FP$\downarrow$ & FN$\downarrow$ & IDSw$\downarrow$\\
  \hline
  \specialrule{0em}{1pt}{0pt}
  \multirow{7}{*}{\rotatebox{90}{MOT17}}
  &FRT~\cite{t_wobnw} &  53.5 &78.0& 52.3 & 19.5 & 36.6 & 12201 & 248047& 2072 \\
  &FRT+DHN~\cite{deepmot} &53.7&77.2&53.8&19.4&36.6&11731&247447&\textbf{1947}\\
  &FRT+BDL~\cite{RNN_FC} &54.9&75.5&53.8&20.2&35.8&11276&240981&2157\\
  &FRT+DCNN~\cite{Deep_lap} &56.3&76.8&55.0&21.2&\textbf{35.2}&9099&235161&2099\\
  &FRT+SD~\cite{sinkhorn_dis} &55.4&77.1&54.6&20.8&35.8&10144&239470&2050\\
  &FRT+RM~\cite{DMMNet} &\textbf{56.4}&\textbf{78.9}&55.3&\textbf{21.3}&35.3&8771&235400&1992\\
  &FRT+SH~\cite{sinkhorn} &56.3&78.8&\textbf{55.4}&21.2&35.3&\textbf{8745}&235612&1996\\
  \cline{2-10}
  \specialrule{0em}{1pt}{0pt}
  &FRT+GLAN &\textbf{56.4}&\textbf{78.9}&55.3&\textbf{21.3}&\textbf{35.2}&8836&\textbf{235156}&1977\\
\hline
  %
\hline
\end{tabular}
\end{spacing}
\end{table*}

To validate the effectiveness of our framework on real scenarios, we employ the multi-object tracking (MOT) problem as the task and embed our GLAN into MOT architecture as an assignment module. In the following, we will introduce the experimental settings and report the experimental results in detail.

MOT can be tackled by exploiting the regression head of a detector to perform temporal alignment of object bounding boxes~\cite{t_wobnw}. Based on the tracking pipeline in~\cite{t_wobnw} which adopts the Faster-RCNN~\cite{FRCNN} as a detector, Xu \emph{et al.}~\cite{deepmot} embed the differentiable DHN into the MOT architecture to train the detector in an end-to-end manner. Similarly, we also combine the differentiable assignment solvers with the tracker in~\cite{t_wobnw} to verify their effectiveness in MOT tasks. Here, we exclude the DFC~\cite{Deep_lap} because it is inflexible to cope with cost matrices with varying sizes. For simplicity, we name the tracker in~\cite{t_wobnw} FRT (\emph{Faster RCNN Tracker}).

In training, the matching between predicted bounding boxes and ground truth is viewed as an assignment problem, and the cost matrix is constructed according to center distance and \emph{Intersection-over-Union} (IOU) between tracks and bounding boxes. Furthermore, we employ the soft MOTA and MOTP (please refer to~\cite{deepmot} for the details of the construction of them) to guide the learning process. In the case of MOT, there can be different numbers of tracks and ground-truth objects, and the objects in the previous frame can be missed in the current frame or a new object can exist in the current frame. Therefore, in training, we remove the constraint loss to prevent the un-assigned track or object from having a high assignment score.
In evaluation, following the tracking pipeline in~\cite{t_wobnw}, only the trained detector is used to perform tracking and the assignment solvers are not employed to facilitate the tracking process.


The performance of each tested method on MOT benchmark MOT17~\cite{mot16} is evaluated by standard MOT metrics 
and is reported in Table~\ref{table:MOT_all}. It is observed that trained along with differentiable assignment solvers, the performance of the tracker has been improved in most metrics.
Specifically, the learnable-free differentiable methods, RM~\cite{DMMNet}, SD~\cite{sinkhorn_dis} and SH~\cite{sinkhorn}, solve the assignment problem directly on the cost matrix, which allows the relationships between each pair of tracks and bounding boxes to be taken into consideration. Therefore, the trackers trained with these solvers have been improved on almost metrics. Despite the improvements achieved by learning-based baselines, there are still several limitations to enhancing the tracker. In detail, the BDL~\cite{RNN_FC} splits the assignment problem into several classification sub-tasks, but the relationships between these sub-tasks are not involved in solving the original assignment problems. The DHN~\cite{deepmot} transforms the cost matrix to a sequential pattern and passes it to an RNN-based model to extract discriminative representations from the global scope. However, it is unsuitable to solve the time-independent problem with the RNN-based model. Besides, the DCNN~\cite{Deep_lap} regards the cost matrix as an image and employs stacked CNN layers to obtain the assignment prediction. But the predicted assignments always reach the local optima when the size of problems exceeds the receptive fields.

In virtue of the structure of the bipartite graph constructed from the cost matrix, the proposed framework GLAN can well capture the global inductive information through message passing, and the tracker trained with GLAN achieves the best results in terms of most metrics. Specifically, in MOT17, both our model and RM~\cite{DMMNet} achieve the best improvement in terms of MOTA ($+2.9\%$), MOTP ($+0.9\%$) and MT ($+1.8\%$) over FRT~\cite{t_wobnw}.
Besides, the tracker combined with our GLAN also obtains the best results in terms of ML and FN, which demonstrates the effectiveness and adaptability of our method for MOT task.

\section{Conclusion}
In this paper, we propose a novel learning framework to improve a differentiable solver for the linear assignment problem. We first convert the problem of making a linear assignment from a cost matrix to the problem of edge selection from a constructed bipartite graph. In order to solve the edge selection task, we propose a graph network to form structured representations for each edge by message passing and predict their labels. Experimental results on synthetic datasets reveal that with low computational complexity, the proposed method outperforms other state-of-the-art baseline solvers on assignment precision. Moreover, as a trainable layer in MOT architecture, the proposed assignment solver can best boost the tracking performance than other differentiable approaches.

\section*{Acknowledgment}
This work is supported by the National Nature Science Foundation of China (Nos. 62076021, 62072027 and 61872032) and the Beijing Municipal Natural Science Foundation (Nos. 4202060, 4202057 and 4212041).
\section*{References}

\bibliography{mybibfile}

\end{document}